\documentclass[10pt,twocolumn,letterpaper]{article}
\usepackage{booktabs, multirow, threeparttable}

\usepackage[T1]{fontenc} 
\usepackage{bm}
\usepackage{overpic}
\usepackage{tabularx}

\usepackage{cvpr}              %

\definecolor{cvprblue}{rgb}{0.21,0.49,0.74}
\usepackage[pagebackref,breaklinks,colorlinks,allcolors=cvprblue]{hyperref}

\def\paperID{1329} %
\def\confName{CVPR}
\def\confYear{2026}

\usepackage{soul}

\title{Animated 3DGS Avatars in Diverse Scenes with Consistent Lighting and Shadows }

\author{Aymen Mir \textsuperscript{1*}  \quad Riza Alp Guler\textsuperscript{2} \quad Jian Wang\textsuperscript{2}  \quad Gerard Pons-Moll \textsuperscript{1} \quad Bing Zhou\textsuperscript{2} \\\\
{\small \textsuperscript{1} Tübingen AI Center, University of Tübingen, Germany} 
\qquad
{\small\textsuperscript{2}Snap Inc.} 
}
\makeatletter

\let\@oldmaketitle\@maketitle%
\renewcommand{\@maketitle}{
	\@oldmaketitle%
	\begin{center}
        \begin{overpic}[abs,unit=1mm,scale=0.275]{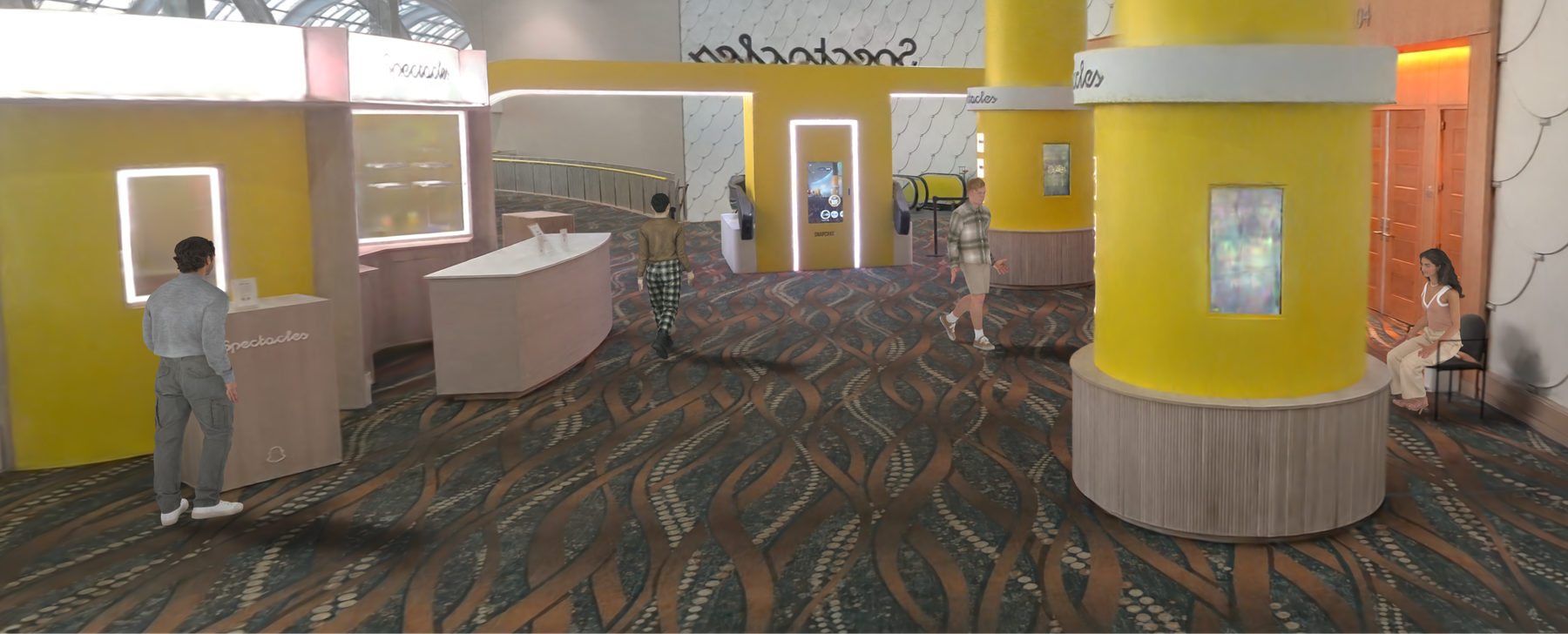}

        \end{overpic}
	\end{center}
    \refstepcounter{figure}\normalfont
    \vspace{-4mm}
{\footnotesize Figure~\thefigure. Our method renders consistent lighting and soft shadows for animated 3DGS avatars interacting with 3DGS scenes. Avatars both cast shadows onto the scene and receive scene illumination via SH‑based relighting, yielding coherent compositions across diverse environments.
}
\label{fig:teaser}

\vspace{1mm}
}
\begin{document}
\maketitle

{\let\thefootnote\relax\footnotetext{*Work done as intern at Snap.}}

\begin{abstract}

We present a method for consistent lighting and shadows when animated 3D Gaussian Splatting (3DGS) avatars interact with 3DGS scenes or with dynamic objects inserted into otherwise static scenes.
Our key contribution is Deep Gaussian Shadow Maps (DGSM)—a modern analogue of the classical shadow mapping algorithm tailored to the volumetric 3DGS representation. Building on the classic deep‑shadow mapping idea, we show that 3DGS admits closed‑form light accumulation along light rays, enabling volumetric shadow computation without meshing. For each estimated light, we tabulate transmittance over concentric radial shells and store them in an octahedral atlases, which modern GPUs can sample in real-time per query to attenuate affected scene Gaussians and thus cast and receive shadows consistently. To relight moving avatars, we approximate the local environment illumination with HDRI probes represented in a spherical‑harmonic (SH) basis and apply a fast per‑Gaussian radiance transfer, avoiding explicit BRDF estimation or offline optimization.
We demonstrate environment consistent lighting for avatars from AvatarX and ActorsHQ, composited into ScanNet++, DL3DV, and SuperSplat scenes, and show interactions with inserted objects. Across single and multi‑avatar settings, DGSM and SH relighting operate fully in the volumetric 3DGS representation, yielding coherent shadows and relighting while avoiding meshing.
\end{abstract}
\vspace{-7mm}
    
\section{Introduction}
Recent work has shown that 3D Gaussian Splatting (3DGS) can represent articulated humans and object–scene interactions at interactive rates \cite{anonymous2026aha, mir2025gaspacho}, often with higher photorealism than mesh-based representations. This opens the possibility of using 3DGS as a representation in content creation and simulation pipelines—for example, virtual production, CG compositing, and simulator construction for robotics and autonomous driving, which have traditionally relied on mesh-based assets. However, when animated Gaussian avatars are inserted into captured Gaussian scenes, a salient gap remains: avatar lighting often fails to match the environment illumination, and shadows are missing, undermining realism.

Two technical hurdles drive this gap. First, Gaussian splats form a volumetric representation that, unlike triangle meshes, lacks explicit watertight surfaces, inside/outside tests, or hard opacity boundaries. Classical raster-space shadow mapping \cite{williams1978casting,reeves1987rendering} and mesh-based light transport therefore do not apply out of the box. Second, even if shadows are handled, relighting a moving avatar so that its shading matches the surrounding scene—under unknown, potentially spatially varying illumination—requires estimating the incident light and transferring it to the avatar in a way that is temporally stable and fast enough for frame-by-frame animation.

We address both challenges with a method (Fig. \ref{fig:teaser}) that computes consistent lighting and shadows for Gaussian avatars interacting with static Gaussian-splat scenes and with dynamic objects inserted into otherwise static scenes. Our central contribution is Deep Gaussian Shadow Maps (DGSM), a modern analogue of deep shadow maps tailored to 3DGS. Inspired by the classic deep-shadow \cite{lokovic2000deepshadow} formulation, we show that Gaussian splatting naturally admits closed-form transmittance and light accumulation along light rays, enabling volumetric deep-shadow computation directly in the Gaussian domain—without meshing, voxelization, or ad-hoc binarization.

Concretely, after estimating scene lights, we build volumetric shadow maps parameterized by concentric spherical shells around each light and store them compactly in an octahedral atlas. GPU kernels allow dynamic sampling of these DGSMs at render time; affected scene Gaussians are attenuated to both cast and receive soft, view-consistent shadows. Because the construction lives in the same space as 3DGS, shadows extend naturally to multiple lights and inserted objects.

To align avatar lighting with the surrounding scene, we approximate local illumination using an HDRI environment probe represented in a real spherical–harmonic (SH) basis. We render a cubemap at the avatar location, and fit SH coefficients via a weighted least–squares solve; the resulting SH compactly encodes incident radiance. Building on standard SH lighting tools~\cite{Mahajan2006_SphericalHarmonicIdentities} and recent SH–based relighting of Gaussian objects~\cite{boyang2025transplatlightingconsistentcrosssceneobject}, we then perform per–Gaussian radiance transfer: for each Gaussian we contract the target environment with a cosine (or glossy) lobe to obtain a per–channel scale, which modulates the Gaussian’s color so its appearance matches the scene’s illumination. The formulation yields a fast, closed–form per–frame update of the SH probe and transfer scales, supporting avatar motion and scene edits without explicit BRDF estimation, inverse rendering, or offline optimization

We validate the approach in three representative settings: (1) single-avatar animation in 3DGS scenes, (2) avatar–object interaction with dynamic props inserted into static scenes, and (3) multi-avatar motion. We demonstrate visually coherent shadows and relighting for animated avatars from AvatarX \cite{zheng2023avatarrex} and ActorsHQ \cite{isik2023humanrf}, interacting with objects from NeuralDome \cite{zhang2023neuraldome}, and composited into scenes from ScanNet++ \cite{yeshwanthliu2023scannetpp}, DL3DV \cite{ling2024dl3dv}, and the SuperSplat \cite{supersplat} library. Across these scenarios, our method operates directly on the volumetric representation, avoiding meshing while delivering  lighting interactions between dynamic avatars and static Gaussian environments.

In summary, this paper offers:
\begin{itemize}
    \item \textbf{Deep Gaussian Shadow Maps (DGSM).} A volumetric deep-shadow formulation for Gaussian splats, with closed-form light accumulation and efficient octahedral-atlas storage for fast sampling.

    \item \textbf{Fast avatar relighting via SH HDRI probes.} A per-frame, per-Gaussian SH transfer that approximates local environment lighting without explicit BRDFs or meshing.

    \item \textbf{Coherent lighting for dynamic 3DGS scenes.} An integrated pipeline that enables avatars and inserted objects to cast shadows and exhibit scene‑matched lighting, validated across ScanNet++, DL3DV, and SuperSplat scenes with AvatarX/ActorsHQ avatars.
\end{itemize}

\section{Related Work}

\textbf{Neural Rendering}
Since the release of NeRF \cite{mildenhall2020nerf}, the area of neural rendering has advanced substantially \cite{nerf_review}. Nevertheless, NeRF remains computationally demanding, and even with a series of accelerations \cite{mueller2022instant, barron2022mip, barron2023zipnerf, nerfstudio}, its overall cost is still high. 3DGS \cite{kerbl2023gaussians} mitigates this by representing a scene with explicit 3D Gaussian primitives, extending earlier point-based ideas \cite{lassner2021pulsar}, and rasterizing them into images using splatting \cite{westover1991phdsplatting}. 
Originally intended for static scenes, 3DGS has since been adapted to dynamic settings \cite{shaw2023swags, luiten2023dynamic, Wu2024CVPR, lee2024ex4dgs, li2023spacetime}, SLAM-style reconstruction \cite{keetha2024splatam}, mesh extraction \cite{Huang2DGS2024, guedon2023sugar}, and sparse-view NVS \cite{mihajlovic2024SplatFields}. While \cite{bolanos2024gsc} supports shadow casting for multi-Gaussian characters, it is not tailored to the 3DGS scene representation and does not handle shadows being \emph{received} in 3D Gaussian scenes. Ray-traced and self-shadowing variants also exist \cite{Byrski2025RaySplats,bi2024rgs}; however, unlike our work, they do not target a general shadow representation for 3DGS that allows shadows to be cast onto 3DGS scenes. Neural representations have moreover been explored for relighting and scene editing \cite{liang2023gs, Gao2023Relightable3DGaussians, chen2023gaussianeditor}, but in contrast to our focus, these works do not address dynamic 3DGS avatars within 3DGS scenes.

\textbf{Human Reconstruction and Neural Rendering}
Mesh templates \cite{SMPL-X:2019, smpl2015loper} are widely used to recover 3D human shape and pose from images or video \cite{Bogo2016keepitsmpl, kanazawaHMR18}, but they do not yield photorealistic renderings. Works such as \cite{alldieck2018video, alldieck19cvpr} reconstruct re-posable avatars from monocular RGB, yet their template-mesh foundation similarly limits photorealism. Implicit representations \cite{mescheder2019occupancy, park2019deepsdf} have been employed to reconstruct detailed clothed humans \cite{chen2021snarf, alldieck2021imghum, saito2020pifuhd, he2021arch++, huang2020arch, deng2020nasa}; however, they also struggle with photorealistic rendering and are often not readily re-posable. A number of methods \cite{peng2021animatable, guo2023vid2avatar, weng_humannerf_2022_cvpr, jian2022neuman, haberman2023hdhumans, heminggaberman2024trihuman, li2022tava, liu2021neural, xu2021hnerf, iqbal2023rana} build controllable NeRFs that produce photorealistic humans from videos, but unlike us, they do not model human–scene interactions. 
With 3DGS, several recent works \cite{ kocabas2023hugs, qian20233dgsavatar, moreau2024human, abdal2023gaussian, zielonka2023drivable, moon2024exavatar, li2024animatablegaussians, pang2024ash, lei2023gart,
qian20233dgsavatar, xu2024gaussian, junkawitsch2025eva} construct controllable human or facial avatars; however, in contrast to our approach, they likewise do not capture human–scene interactions. Some works extend 3DGS to model humans together with their environment \cite{xue2024hsr, tomie, mir2025gaspacho}, but unlike our work, they do not focus on animating humans within 3D scenes nor on consistent lighting and shadows.

\textbf{Humans and Scenes}
Human–scene interaction has been a long-standing topic in vision and graphics. Early efforts \cite{fouhey2014people, wang2017binge, gupta20113d} infer affordances and interactions from monocular RGB. The availability of large-scale HSI datasets \cite{Hassan2021-gb, mir20hps, hassan2019prox, savva2016pigraphs, taheri2020grab, bhatnagar22behave, jiang2024scaling, cheng2023dnarendering, zhang2022couch, mahmood19amass, BABEL:CVPR:2021, Guo_2022_CVPR_humanml3d} has driven progress in joint 3D reconstruction of human–object interactions \cite{xie2022chore, xie2023vistracker, xie2024template, zhang2020phosa} and in methods for object-conditioned, controllable human motion \cite{zhang2022couch, starke19neural, hassan21cvpr, diller2023cghoi, zhang2022wanderings, Zhao:DartControl:2025, zhao2023dimos}. These approaches typically use mesh representations for both humans and scenes and thus inherit mesh-related limitations, while our method supports photorealistic renderings of humans and environments with lighting and shadow consistency.

\textbf{Human Relighting} Recent portrait relighting focuses largely on image-space learning, including encoder–decoder approaches trained on light-stage data and physics-guided decompositions (\cite{Sun2019PortraitRelighting,Kanamori2018RelightingHumans,Pandey2021TotalRelighting,Ji2022GeometryAware,Kim2024SwitchLight}) as well as diffusion-based formulations for faces (\cite{He2024DifFRelight,Ho2020DDPM,Rombach2022LDM,Song2021SDE,Song2021DDIM}). In parallel, physically based and inverse-rendering methods recover materials, lighting and geometry and enable relightable human capture and avatars (\cite{Hasselgren2022nvdiffrecmc,Jin2023TensoIR,Guo2019Relightables,ChenLiu2022Relighting4D,Saito2024RGCA,Iwase2023RelightableHands,Chen2024URHand,Zhang2021NVPRelighting,Sarkar2023LitNeRF,WangARXIV2025}). While these works often study relighting in isolation, our goal is different: we transfer lighting from a surrounding 3DGS environment onto the human subject with explicit attention to shadow formation and consistency.

\begin{figure*}
    \centering
    \includegraphics[width=1.01\linewidth]{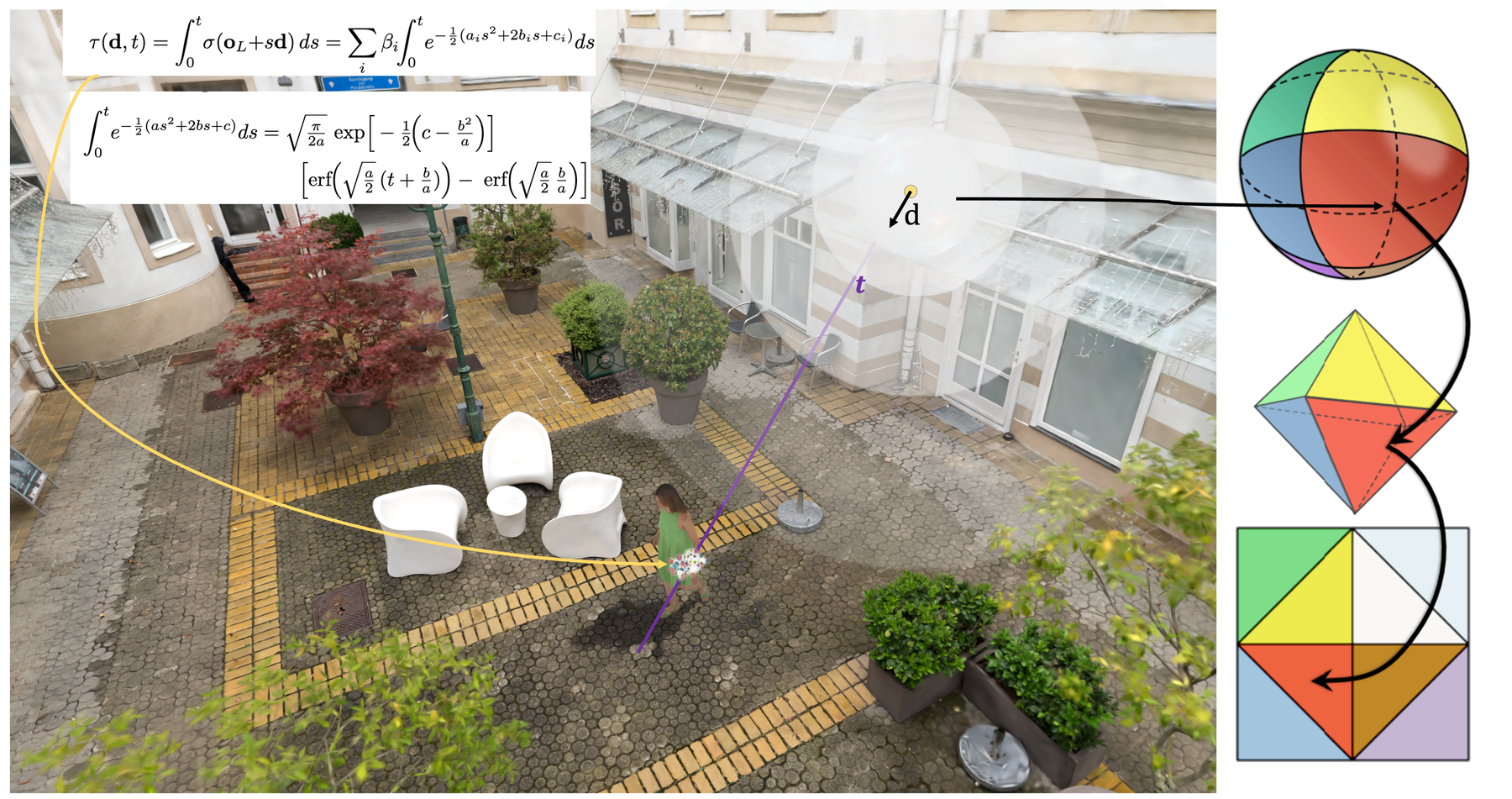}
    \caption{\textbf{Deep Gaussian Shadow Maps:} For concentric spheres radiating out from light source, we build DGSMs by computing the light absorption by inserted Avatar/Object Gaussian at each radial distance from the light source. An octahedral map (right) takes a 3D unit vector $\textbf{d}$ and maps it to a 2D location in the atlas - of fixed dimension $H \times W$. Each of the absorption values in the concentric spheres is mapped to its own 2D octahedral map. The radial distances of the spheres are chunked into K discrete bins  - along the radial direction $\textbf{t}$ - and stored in octahedral atlases. This creates a volumetric shadow map of fixed dimension $K \times H \times W$ which can be sampled to cast shadows on the Scene Gaussians.}
    \label{fig:method_figurre}
\end{figure*}

\section{Method}
\label{sec:method}

We assume dynamic human Gaussians $\mathcal{G}_h$ and optional dynamic object Gaussians $\mathcal{G}_O$ are placed in the scene Gaussians $\mathcal{G}_s$. To this end we use existing pipelines that generate such dynamic human or object Gaussians interacting with scene Gaussians. Our goals are two fold: 1) relight $\mathcal{G}_h$ to match local scene illumination; 2) enable $\mathcal{G}_h$ and $\mathcal{G}_O$ to cast and the scene to receive soft shadows. We first estimates k dominant light sources (Sec. \ref{subsec:light}) for shadow casting. We then build Deep Gaussian Shadow Maps using the estimated light sources (Sec. \ref{subsec:dsm_build}) and sample the shadow map on the scene Gaussians (Sec. \ref{subsec:dsm_sample}) to cast shadows. To change the color of the human Gaussians we first estimate an approximate HDRI environment map (Sec. \ref{subsec:hdri_build}) and transfer the environment properties onto the inserted Avatar/Object Gaussians (Sec. \ref{subsec:hdri_transfer}). 

\subsection{High Intensity Light Source Estimation}
\label{subsec:light}

Our method estimates a compact set of k point light sources from a Gaussian scene representation. We first center an ROI around the character’s alpha-weighted centroid and evaluate per-Gaussian color using the provided spherical-harmonics (SH) coefficients from a small set of viewpoints near the character. From these multi-view SH evaluations we derive simple photometric cues—mean/max luminance, an angular-stability term, and a DC-dominance prior—and softly clip extreme outliers. A base intensity score combines these cues without any bias toward larger Gaussians, and we keep only the high-score tail for efficiency.

Within this set, we promote compact emitters by measuring local contrast: each candidate’s score is compared against the average score in a small spatial neighborhood, yielding a peakness measure that naturally downweights broad emitters. Finally, we select k lights via greedy distance-based NMS suppression—no clustering or plane fitting required. The selected Gaussian centers define light positions; their intensities come from the base score; and colors are obtained by evaluating the SH toward the character. This yields a succinct, robust set of point sources driven purely by photometric evidence and local contrast.

\subsection{Deep Gaussian Shadow Maps - Build}
\label{subsec:dsm_build}

We render shadows from a light (Fig. \ref{fig:method_figurre}) for scenes and characters or inserted objects represented by anisotropic 3D Gaussians. The $i$‑th Avatar/Object Gaussian has mean $\boldsymbol{\mu}_i\!\in\!\mathbb{R}^3$, covariance $\boldsymbol{\Sigma}_i\!\in\!\mathbb{R}^{3\times3}$ , precision $\mathbf{A}_i=\boldsymbol{\Sigma}_i^{-1}$, and per‑Gaussian opacity $\alpha_i\!\in\!(0,1)$. A world point $\mathbf{x}$ is seen from the light at $\mathbf{o}_L$ along the ray
$\mathbf{r}(t)=\mathbf{o}_L+t\,\mathbf{d}$ with unit direction
$\mathbf{d}=(\mathbf{x}-\mathbf{o}_L)/\|\mathbf{x}-\mathbf{o}_L\|$, $t\!\ge\!0$.

\noindent\textbf{Volumetric visibility in closed form.}
We first model the light absorption field as a Gaussian mixture with an explicit relationship, defined below, between the absorption coefficient $\beta$ and Gaussian opacity $\alpha$
\begin{equation}
\sigma(\mathbf{x})=\sum_i \beta_i\,
\exp\!\Big(-\tfrac{1}{2}\,(\mathbf{x}-\boldsymbol{\mu}_i)^\top
\mathbf{A}_i(\mathbf{x}-\boldsymbol{\mu}_i)\Big).
\end{equation}
The optical depth to distance $t$ factorizes per Gaussian:
\begin{equation}
\label{eq:tau_sum}
\tau(\mathbf{d},t)
=\int_{0}^{t}\!\sigma(\mathbf{o}_L+s\mathbf{d})\,ds
=\sum_i \beta_i \!\!\int_{0}^{t}\!
e^{-\frac{1}{2}(a_i s^2+2b_i s+c_i)} ds, 
\end{equation}
$a_i=\mathbf{d}^\top\mathbf{A}_i\mathbf{d},
b_i=\mathbf{d}^\top\mathbf{A}_i(\mathbf{o}_L-\boldsymbol{\mu}_i),
c_i=(\mathbf{o}_L-\boldsymbol{\mu}_i)^\top\mathbf{A}_i(\mathbf{o}_L-\boldsymbol{\mu}_i)$
Each term admits the standard error‑function primitive:
\begin{align}
\label{eq:closed_form}
\int_{0}^{t}\!e^{-\frac{1}{2}(a s^2+2 b s+c)} ds
&= \sqrt{\tfrac{\pi}{2a}}\,\exp\!\Big[-\tfrac{1}{2}\!\Big(c-\tfrac{b^2}{a}\Big)\Big]\,
\nonumber\\[-2pt]
&\qquad
\Big[
\mathrm{erf}\!\Big(\sqrt{\tfrac{a}{2}}\,(t+\tfrac{b}{a})\Big)  -\;
\mathrm{erf}\!\Big(\sqrt{\tfrac{a}{2}}\,\tfrac{b}{a}\Big)
\Big].
\end{align}
We use this to calculate a mapping between direction $\textbf{d}$ and distance $t$ from light source - Transmittance (visibility) as
\begin{equation}
\label{eq:T_of_tau}
\mathcal{T}(\mathbf{d},t)=\exp\!\big[-\tau(\mathbf{d},t)\big].
\end{equation}

\noindent\textbf{3DGS Opacity$\rightarrow$absorption calibration.}
Let $\tau_i^\star=-\ln(1-\alpha_i)$ be the optical depth implied by the 3DGS image formation model. Because Eq.\eqref{eq:closed_form} introduces a direction‑dependent $1/\sqrt{a_i}$ factor, we set the absorption amplitude using a direction‑agnostic proxy for $\sqrt{a_i}$ to stabilize shadow strength across scales and anisotropies:
\begin{align}
\label{eq:beta_trace}
\beta_i
&=\kappa\,\tau_i^\star\,
\frac{\sqrt{\mathrm{tr}(\mathbf{A}_i)/3}}{\sqrt{2\pi}},
\end{align}
where $\kappa$ is a global strength knob. We evaluate other design choices for $\beta$ (See Experiments - Sec \ref{sec:exp}) and find \eqref{eq:beta_trace} preserves fine shadow detail more consistently.

\noindent\textbf{Directional parameterization, discretization, and storage.}
Relative to evaluating spherical functions on the fly, we map the function values to an octahedral atlas (Fig. \ref{fig:method_figurre}). The atlas turns a spherical function into a single contiguous 2D texture, enabling precomputation, compact storage, and fast, vectorized sampling on GPUs. Compared to cubemaps, it avoids inter‑face seams and exhibits more uniform angular distortion; we confirm both effects in ablations.

We tabulate $\mathcal{T}$ over directions and distance for $O(1)$ sampling at render time. Directions on $\mathbb{S}^2$ are encoded with an octahedral atlas, i.e., a parameterization 
$\psi : \mathbb{S}^2 \to D \subset \mathbb{R}^2$, which maps the sphere onto a 2D domain. Given a unit vector $\mathbf{d}=(x,y,z)$, we encode $\mathbf{q}=\frac{\mathbf{d}}{|x|+|y|+|z|}$
\begin{align}
(u,v)=
\begin{cases}
(q_x,q_y), & q_z\ge 0,\\[2pt]
(\mathrm{sgn}(q_x)\,[1-|q_y|],\;\mathrm{sgn}(q_y)\,[1-|q_x|]), & q_z<0,
\end{cases}
\end{align}
then rescale $(u,v)\!\in\![-1,1]^2$ to an $H\times W$ texture grid (pixel centers). The inverse decodes by undoing the fold, setting $z=1-|u|-|v|$, and normalizing. Distance is discretized into $K$ radial bins with centers $t_k=(k+\tfrac{1}{2})\,t_{\max}/K$. We thus store a 3D table
$\mathcal{T}[u,v,k]\approx \exp[-\tau(\mathbf{d}(u,v),t_k)]$ and sample it with GPU trilinear interpolation.

\noindent\textbf{Receiver‑driven region of interest (ROI).}
Computing $\mathcal{T}$ densely over all $(u,v,k)$ is unnecessary and expensive. Let $\mathbf{c}$ be the $\alpha$‑weighted centroid of Avatar Gaussians and $[z_{\min},z_{\max}]$ robust height bounds. We restrict receivers (where the integral is computed) to
\(
\mathcal{B}=\{\mathbf{x}:\|(\mathbf{x}-\mathbf{c})_{xy}\|_\infty\le R,\; z_{\min}\le x_z\le z_{\max}\}
\)
with a known radius $R$ (e.g., $2$\,m). For scene Gaussians whose centers lie in $\mathcal{B}$, we project their light rays into atlas pixels, collect the unique set $\mathcal{P}$, and infer a tight radial range $k\!\in\![k_{\min},k_{\max}]$ from their light‑space distances. We initialize $\mathcal{T}\!\equiv\!1$ and \emph{only} accumulate optical depth on the voxel slab $\mathcal{R}=\mathcal{P}\times\{k_{\min},\ldots,k_{\max}\}$; outside $\mathcal{R}$ the table remains $\mathcal{T}\!=\!1$ by construction.

\noindent\textbf{Occluder culling via light‑space footprints.}
Even inside $\mathcal{R}$, most inserted Avatar Gaussians do not affect a given atlas pixel. Analogous to tile‑based culling in 3D Gaussian Splatting \cite{kerbl2023gaussians}, we compute for each occluder a conservative ellipse on the light’s tangent plane and ignore non‑overlapping occluders. If $\mathbf{d}_i$ is the unit vector from $\mathbf{o}_L$ to $\boldsymbol{\mu}_i$ and $(\mathbf{u}_i,\mathbf{v}_i)$ is an orthonormal basis of the plane orthogonal to $\mathbf{d}_i$, the projected covariance is
\(
\boldsymbol{\Sigma}_{\perp,i}=
[\mathbf{u}_i\ \mathbf{v}_i]^\top
\boldsymbol{\Sigma}_i
[\mathbf{u}_i\ \mathbf{v}_i].
\)
Let its eigenvalues be $\lambda_{1,i}\!\ge\!\lambda_{2,i}$. Angular standard deviations are
$\sigma_{j,i}=\sqrt{\lambda_{j,i}}/\|\boldsymbol{\mu}_i-\mathbf{o}_L\|$; with pixels‑per‑radian
$\rho\!\approx\!(H{+}W)/(2\pi)$, a $k_\sigma$‑rule ellipse has radii
$p_{j,i}=k_\sigma\,\sigma_{j,i}\,\rho$ in atlas pixels. We bucket occluders into $8{\times}8$ atlas tiles using these rectangles. For each ROI pixel we then gather only the few occluders in its buckets before evaluating \eqref{eq:closed_form}, reducing the per‑pixel sum from $O(N_{\text{occ}})$ to $O(\bar{M})$ with $\bar{M}\!\ll\!N_{\text{occ}}$.

\noindent\textbf{Relation to 3DGS culling.}
3DGS rasterization computes a screen‑space covariance and uses its eigenvalues to bound each Gaussian’s footprint, updating only the covered tiles \cite{kerbl2023gaussians}. Our light‑space culling is the exact analogue: we compute a covariance on the plane orthogonal to the light ray, convert it to pixel radii on the directional atlas, and restrict accumulation to overlapping tiles. Combined with the receiver‑driven ROI (which prunes the \emph{domain} of the atlas itself), this yields two complementary reductions: fewer voxels to update and far fewer occluders per voxel.

\subsection{Deep Gaussian Shadow Maps - Sampling}
\label{subsec:dsm_sample}

At render time, each scene Gaussian center $\mathbf{x}_s$ fetches
$\mathcal{T}\!\left[\psi(\mathbf{d}_s),\,t(\mathbf{x}_s)\right]$
with $\mathbf{d}_s=(\mathbf{x}_s-\mathbf{o}_L)/\|\mathbf{x}_s-\mathbf{o}_L\|$ and $t(\mathbf{x}_s)=\|\mathbf{x}_s-\mathbf{o}_L\|$, (with $\mathbf{o}_L$ denoting a light source) via trilinear interpolation. We multiply the direct term of its color by this transmittance and then render the modulated scene and the Avatar using the standard 3DGS splatting pipeline.

In practice we estimate per-receiver transmittance by sampling a small footprint around each Gaussian center and averaging deep-shadow lookups over those points. By default we use Monte Carlo sampling. In ablations (Sec. \ref{sec:exp}) we also experiment with a deterministic 7-point stencil that places fixed offsets in the principal-axes frame and aggregates them with normalized kernel weights.

\subsection{Approximate HDRI map - Build}
\label{subsec:hdri_build}

We construct an approximate environment by rendering the 3DGS scene on the six \(90^\circ\) cube faces at Avatar location, producing RGB samples \(Y\in\mathbb{R}^{N\times 3}\), unit directions \(D\in\mathbb{R}^{N\times 3}\), and per-pixel solid-angle weights \(w\in\mathbb{R}^N\) that correct cubemap area distortion. On these directions we evaluate the real spherical-harmonic (SH) basis up to degree \(d\), forming \(B\in\mathbb{R}^{N\times K}\) with \(K=(d+1)^2\). The SH coefficients per color channel, \(A\in\mathbb{R}^{K\times 3}\), follow from a weighted ridge least-squares problem
\begin{equation}
\min_{A\in\mathbb{R}^{K\times 3}}\;\bigl\|W^{1/2}(BA-Y)\bigr\|_F^{2}+\lambda\|A\|_F^{2},
\end{equation}
so $(B^{\!\top}WB+\lambda I)A=B^{\!\top}WY$ 

with \(W=\mathrm{diag}(w)\). The solution, reshaped as \(\mathrm{SH}\in\mathbb{R}^{3\times K}\), compactly represents the environment and can be evaluated at arbitrary directions via
\begin{equation}
\hat L(\omega) \;=\; B(\omega)\,A \;\in\; \mathbb{R}^{1\times 3}.
\end{equation}

\subsection{Transfer lighting using approx HDRI }
\label{subsec:hdri_transfer}

To transfer lighting onto a Avatar’s Gaussian elements, we sample a latitude–longitude grid \(\{\omega_j\}_{j=1}^{M}\) with quadrature weights \(w_j\propto \sin\theta_j\) and recover radiance \(L(\omega_j)\) from the fitted SH. For a Gaussian with unit normal \(\mathbf n\), a cosine lobe
\begin{equation}
S(\omega,\mathbf n) \;=\; \max\!\bigl(0,\langle\omega,\mathbf n\rangle\bigr)^{q}
\end{equation}
(Lambertian when \(q=1\)) aggregates incident light. The per-channel lighting scale is the normalized, weighted contraction
\begin{equation}
s_c(\mathbf n)=\operatorname{clip}_{[0,t_{\max}]}\!\left(
\frac{\sum_{j=1}^{M} w_j\,L_c(\omega_j)\,S(\omega_j,\mathbf n)}{\sum_{j=1}^{M} w_j\,S(\omega_j,\mathbf n)+\varepsilon}
\right), 
\end{equation}
where $c\in\{r,g,b\}$ which is robust to exposure and sampling density. The relit color is then $\mathbf c'=\max\!\bigl(\mathbf 0,\;\gamma\,\mathbf c\odot \mathbf s(\mathbf n)\bigr)$ with original color \(\mathbf c\), global intensity \(\gamma\), and elementwise product \(\odot\), yielding efficient image-based lighting consistent with the estimated environment.

\section{Experiments}
\label{sec:exp}

\subsection{Lighting Consistency (no GT)}
\label{sec:metrics-lighting-short}

\paragraph{Setup.}
Let \(\mathcal{S}\) denote the set of scene Gaussians and \(\mathcal{A}\) the set of avatar Gaussians (either \texttt{orig} or \texttt{relit}); \(|\mathcal{A}|\) is the number of avatar Gaussians used. Each Gaussian \(i\) has a center \(\mathbf{x}_i\in\mathbb{R}^3\) and a unit pseudo-normal \(\hat{\mathbf{n}}_i\in\mathbb{S}^2\) estimated by alpha-/distance-weighted local PCA on Gaussian centers (or from the smallest ellipsoid axis when available). Let \(\mathbf{I}_i\in\mathbb{R}^3\) be the per-Gaussian RGB intensity and \(Y(\mathbf{I}_i)\in\mathbb{R}\) its luminance (CIE \(Y\)). For eval, we model irradiance with real spherical harmonics (SH) of order \(L{=}3\): \(E(\hat{\mathbf{n}};\mathbf{c})=\sum_{l=0}^{L}\sum_{m=-l}^{l} c_{lm}\,Y_{lm}(\hat{\mathbf{n}})\), where \(Y_{lm}(\cdot)\) are SH basis functions and \(\mathbf{c}=[c_{lm}]_{l,m}\in\mathbb{R}^{(L+1)^2}\) are SH coefficients. A nearby-scene neighborhood is \(\mathcal{N}=\{\,i\in\mathcal{S}\mid \|\mathbf{x}_i-\mathbf{x}_{\text{avatar}}\|\le r\,\}\), where \(r\in[1,2]\) m and \(\mathbf{x}_{\text{avatar}}\) is a reference avatar position (e.g., its centroid). 

\paragraph{(1) Probe--Avatar Agreement in Luminance (PAA--Y).}
\textit{Intuition:} the avatar and its surrounding scene should ``see'' the same lighting lobes.
We estimate \(\mathbf{c}^{\text{scene}}\in\mathbb{R}^{(L+1)^2}\) and per-Gaussian scales \(\{\alpha_i\ge 0\}_{i\in\mathcal{N}}\) by factorizing \(Y(\mathbf{I}_i)\approx \alpha_i\,E(\hat{\mathbf{n}}_i;\mathbf{c}^{\text{scene}})\) for \(i\in\mathcal{N}\). Estimate \(\mathbf{c}^{\text{avatar}}\in\mathbb{R}^{(L+1)^2}\) and \(\{\beta_a\ge 0\}_{a\in\mathcal{A}}\) from the avatar via \(Y(\mathbf{I}_a)\approx \beta_a\,E(\hat{\mathbf{n}}_a;\mathbf{c}^{\text{avatar}})\). We define the metric as \(\mathrm{PAA\text{-}Y}=\|\mathbf{c}^{\text{avatar}}-\mathbf{c}^{\text{scene}}\|_1\) and report the improvement as \(\Delta\mathrm{PAA\text{-}Y}\equiv \mathrm{PAA\text{-}Y}_{\text{relit}}-\mathrm{PAA\text{-}Y}_{\text{orig}}\) (lower is better).

\paragraph{(2) Avatar Photometric Fit in Luminance (APF--Y).}
\textit{Intuition:} scene-estimated light should predict avatar shading as a function of surface orientation. Given \(\mathbf{c}^{\text{scene}}\), we predict avatar luminance with a per-frame affine calibration \((s,b)\in\mathbb{R}^2\): \(\hat{Y}_a=s\,E(\hat{\mathbf{n}}_a;\mathbf{c}^{\text{scene}})+b\) for \(a\in\mathcal{A}\). With observed luminance \(Y^{(\cdot)}_a\) for \((\cdot)\in\{\texttt{orig},\texttt{relit}\}\),we define \(\mathrm{APF\text{-}Y}=\frac{1}{|\mathcal{A}|}\sum_{a\in\mathcal{A}}\big|\,Y^{(\cdot)}_a-\hat{Y}_a\,\big|\) and report the improvement as \(\Delta\mathrm{APF\text{-}Y}\equiv \mathrm{APF\text{-}Y}_{\text{relit}}-\mathrm{APF\text{-}Y}_{\text{orig}}\) (lower is better).

\paragraph{(3) Chromaticity Neighborhood Match (NCM--\(ab\)).}
\textit{Intuition:} the avatar’s color cast should match the local scene.
We convert \(\mathbf{I}_a\) and \(\mathbf{I}_i\) to CIE--Lab; keep chromaticities \((a^*_a,b^*_a)\) for \(a\in\mathcal{A}\) and \((a^*_i,b^*_i)\) for \(i\in\mathcal{N}\). Let \(P^{\text{avatar}}_{ab}\) be the empirical distribution of \(\{(a^*_a,b^*_a)\}_{a\in\mathcal{A}}\) and \(P^{\text{scene}}_{ab}\) that of \(\{(a^*_i,b^*_i)\}_{i\in\mathcal{N}}\). We define \(\mathrm{NCM\text{-}ab}=\mathrm{EMD}\!\big(P^{\text{avatar}}_{ab},P^{\text{scene}}_{ab}\big)\), where \(\mathrm{EMD}\) is the 1--Wasserstein distance on \(\mathbb{R}^2\) and  the improvement as \(\Delta\mathrm{NCM\text{-}ab}\equiv \mathrm{NCM\text{-}ab}_{\text{relit}}-\mathrm{NCM\text{-}ab}_{\text{orig}}\) (lower is better).

\paragraph{Protocol.}
We use a neighborhood radius \(r\in[1,2]\) m for \(\mathcal{N}\), apply Huber-loss fitting with \(5\%\) trimming when estimating \(\mathbf{c}^{\text{scene}}\), \(\mathbf{c}^{\text{avatar}}\), and per-Gaussian scales. We set the SH order to \(L{=}3\). For each clip, compute all three metrics for \texttt{orig} and \texttt{relit}, and headline the deltas \(\Delta\) as evidence of increased avatar--scene lighting and color consistency without ground-truth illumination or albedo. In Tab. \ref{tab:lighting-consistency-nice} we report Lighting consistency metrics across 3 scenes.

\subsection{Shadow Map Evaluation (Pseudo-GT)}
\label{sec:shadow-eval}

\textit{Pseudo-GT from meshes.}
In the absence of Ground Truth shadow maps, we use mesh based pseudo GT shadow maps for evaluation. We replace the avatar Gaussians with the posed SMPL mesh and render classical shadow maps under the same lights/cameras. For this evaluation protocol we use ScanNet++ which has both 3DGS and mesh scenes available. Using a transparent (or white) receiver mesh, we obtain a \emph{shadow-only} image which we use as pseudo ground-truth. We denote pseudo GT shadow maps as $S^\dagger(p)$

\textit{Shadow map in Gaussian space.}
We render a receivers-only pass in GS: set all \emph{scene} Gaussian colors to zero and accumulate only shadow strength. For computing shadow map value we use our Deep Gaussian Shadow Map formulation. We denote Gaussian shadow maps as $S(p)$

\textit{Metrics.}
We evaluate in an avatar-centric ROI. Let $M^\dagger(p)=\mathbb{1}[S^\dagger(p)>\tau]$, $M(p)=\mathbb{1}[S(p)>\tau]$ with $\tau{=}0.1$, and $\Omega_s=\{p: M^\dagger(p)=1\}$. We report three pixel-space scores:
\begin{itemize}
\item \textbf{SAE (attenuation error):} $\displaystyle \mathrm{SAE}=\frac{1}{|\Omega_s|}\sum_{p\in\Omega_s}\!\big|S(p)-S^\dagger(p)\big|$ \;(lower is better).
\item \textbf{SM-IoU (shadow matte IoU):} $\displaystyle \mathrm{IoU}=\frac{|M\cap M^\dagger|}{|M\cup M^\dagger|}$ \;(higher is better).
\item \textbf{BF (boundary F-measure):} F-score between the boundaries of $M$ and $M^\dagger$ \;(higher is better).
\end{itemize}

\begin{figure*}
    \centering
    \includegraphics[width=0.99\linewidth]{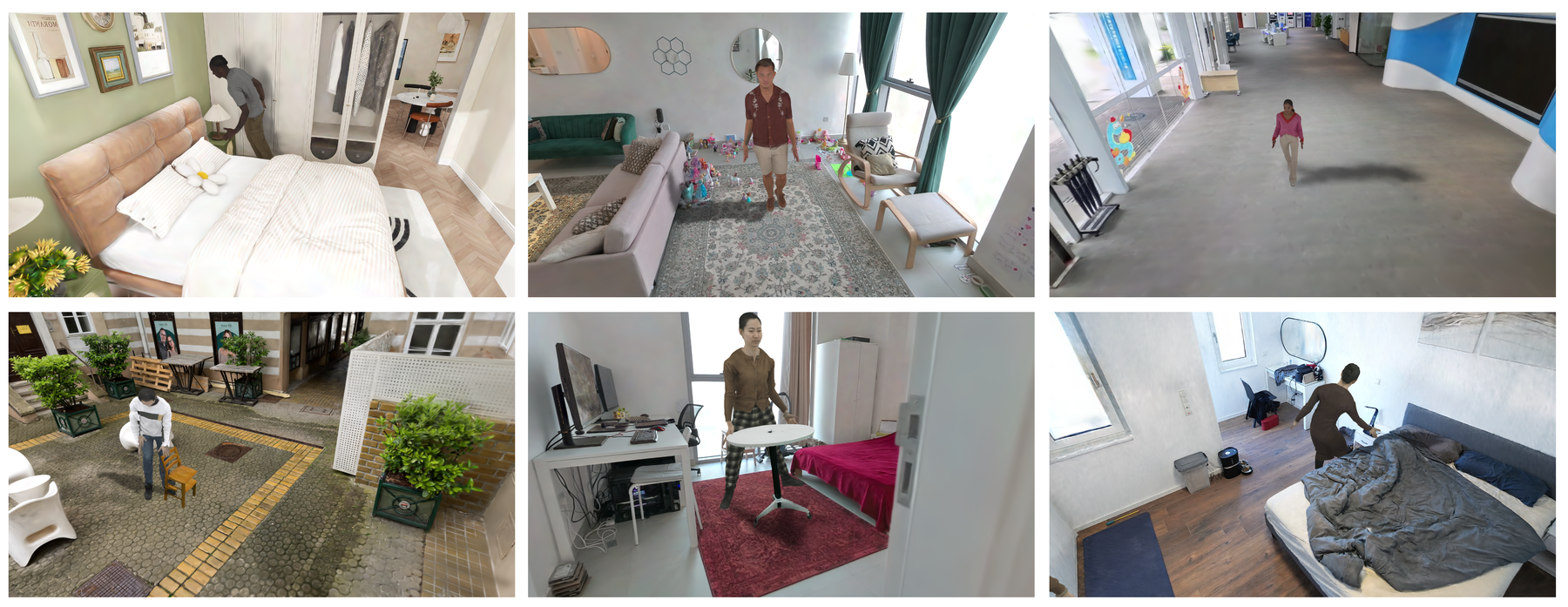}
    \caption{Qualitative Results: Our method casts plausible shadows for various Gaussian Avatars animated alone and with objects.}
    \label{fig:qualitative_results}
\end{figure*}

\begin{figure*}
    \centering
    \includegraphics[width=0.99\linewidth]{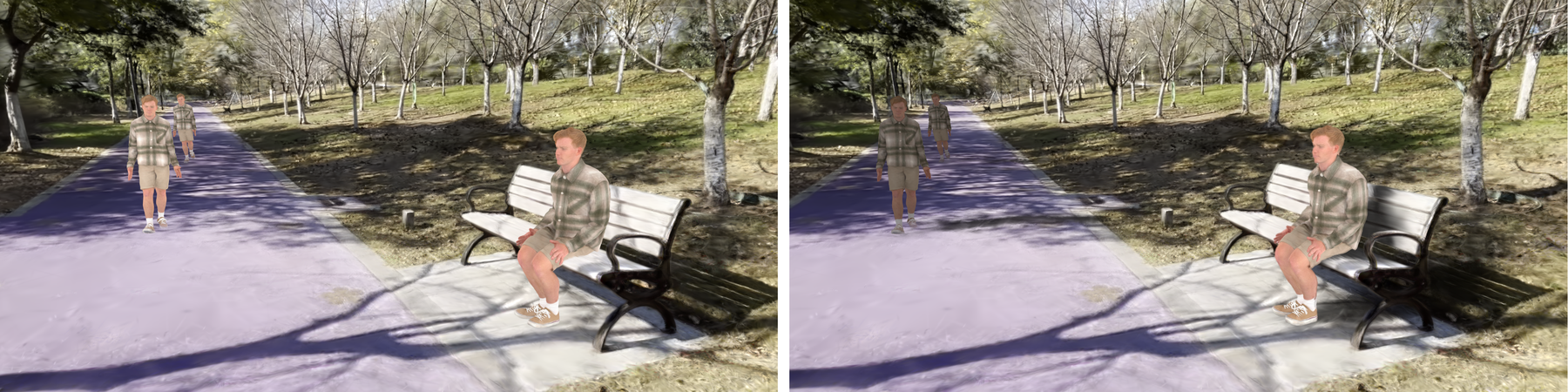}
    \caption{\textbf{Left:} Without our relighting and shadow casting, the animated Avatar in 3D Gaussian scenes does not accurately reflect the lighting effects of the environment, nor does it cast shadows in the scene. The lighting of the Avatar remains uniform throughout. \textbf{Right:} With our SH based relighting and Deep Gaussian Shadow Map Formulation, the Avatar accurately reflects the lighting in the environment and casts accurate shadows.}
    \label{fig:comparison}
\end{figure*}

\subsection{Perceptual Study (Full-Rendering)}
\label{sec:perceptual-short}

We aim to evaluate overall realism and lighting/shadow plausibility. We use \textbf{3} scenes (\textbf{3} clips) per scene, each 3\,s at 24\,fps, 720p; identical cameras/lights. We compare: \textbf{Ours} with \cite{anonymous2026aha} - \textbf{NoRelight+NoShadow} baseline. We pose two questions: which looks more realistic? which better matches scene lighting/shadows?

We ask 12 naive raters with randomized method order and left/right placement to rate the two clips and report aggregate win rates for Realism/Lighting. We report results in Tab. \ref{tab:perceptual-nice}. Fig. \ref{fig:comparison} shows a qualitative example.

\subsection{Ablation: Components of the Shadow-Mapping Pipeline}
\label{sec:ablation-shadow}

\begin{figure*}
    \centering
    \includegraphics[width=0.99\linewidth]{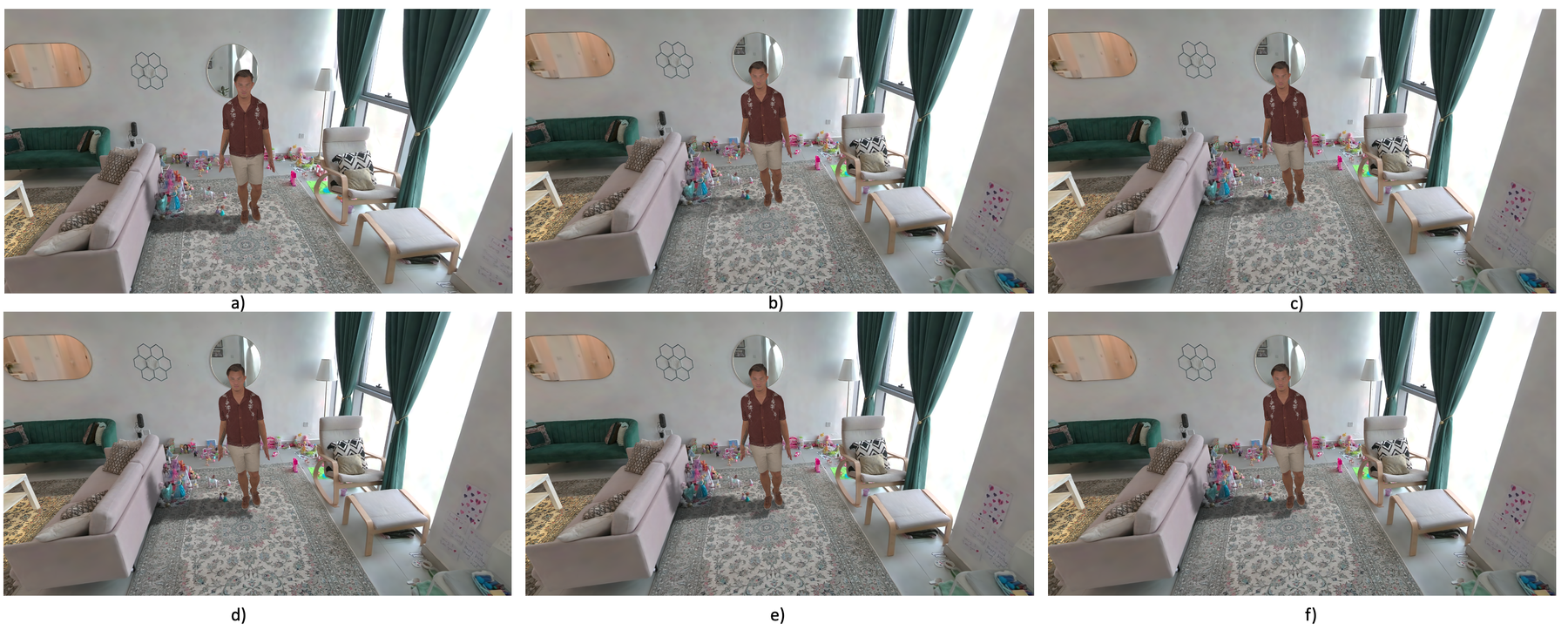}
    \caption{\textbf{Ablation Studies:} Variations of shadow map parameters. (a) Full method; (b) simple opacity-to-absorption; (c) diagonal opacity-to-absorption; (d) cubemaps instead of octahedral maps; (e) center sampling instead of MC; (f) deterministic samplings instead of MC.}
    \vspace{-5mm}
    \label{fig:ablations}
\end{figure*}

\paragraph{(A) Footprint sampling vs.\ center-sample.}
For each receiver Gaussian $g$ (center $\boldsymbol{\mu}_g$, rotation $R_g$, scales $s_g$), we evaluate deep-shadow transmittance either at a \emph{single point} (“center-sample”)
$T_g^{\text{ctr}} \;=\; T\big(\boldsymbol{\mu}_g\big)$
or by \emph{rotation-aware footprint sampling} in principal axes with offsets $\mathbf{z}_i$ (deterministic 7-point stencil when $n\!\ge\!7$, or Monte Carlo sampling): (7 point stencil first defined in \cite{williams2009roofline})

\[
\mathbf{x}_{g,i}=\boldsymbol{\mu}_g + R_g\,(s_g\!\odot\!\mathbf{z}_i),
T_g^{\text{fp}}=\sum_i w_i\,T(\mathbf{x}_{g,i}),\;\; \sum_i w_i=1.
\]
The footprint MC sampling respects the Gaussian’s spatial extent, reducing aliasing/ringing on thin occluders and producing smoother penumbrae. Results in Tab. \ref{tab:ablation-footprint} (Fig. \ref{fig:ablations})

\paragraph{(B) Opacity to absorption ($\alpha\!\to\!\beta$). Eq. \ref{eq:beta_trace}}
With optical depth $\tau=-\log(1-\alpha)$, we test mappings:
\[
\textit{1}:~\beta=\kappa\,\tau;\quad
\textit{2}:~\beta=\tau\;\frac{\sqrt{\operatorname{tr}(A)/3}}{\sqrt{2\pi}};
\]
\[
\textit{3}:~\beta=\frac{\tau}{(2\pi)^{3/2}\sqrt{\det A}};\quad
\textit{4}:~\beta=\frac{\tau}{(2\pi)^{3/2}\,s_x s_y s_z}.
\]
Here $A$ is the covariance in the Gaussian’s local frame (or its diagonal scales $s_x,s_y,s_z$). These control how per-splat opacity distributes to volumetric absorption. We empirically find the mapping 2 that dilutes effect of spatial extent works best. Results in Tab. \ref{tab:ablation-alpha2beta} (Fig. \ref{fig:ablations})

\paragraph{(C) Atlas parameterization (Octahedral vs.\ Cubemap).}
We store per-light deep-shadow fields in an \emph{octahedral} atlas (ours) vs.\ a \emph{cubemap} (6 faces). Octahedral mapping reduces face seams at grazing directions and improves cache coherence; we quantify quality in pixel space against the SMPL pseudo-GT (Sec.~\ref{sec:shadow-eval}). Results in Tab. \ref{tab:ablation-atlas}. (Fig. \ref{fig:ablations})

\paragraph{(D) Culling (timing).}
We cull casters in light space and receivers in scene space around the avatar. In this experiment we compute build time for DGSM across 5 scenes. Disabling ROI culling on an A100 GPU increases  build time from 0.13 s/frame to 29.1 s/frame; disabling light-space culling increases build time from 0.13 s/frame to 17.1 s/frame; thus highlighting the necessity of the two culling mechanisms. 

\subsection{Quantitative Results}
Here we demonstrate that our method works for single Avatar animation in Gaussian scenes, multiple Avatar animation, and for Gaussian Avatars interacting with Dynamic Objects in 3D Gaussian scenes. (Fig. \ref{fig:qualitative_results})

\begin{table}[t]
\centering
\footnotesize
\setlength{\tabcolsep}{3.5pt}
\renewcommand{\arraystretch}{1.15}
\begin{threeparttable}
\caption{Lighting consistency (no GT). Lower is better for PAA--Y / APF--Y / NCM--$ab$. We report Original (non-relit), Relit (ours), and the improvement $\Delta=\text{Orig}-\text{Relit}$.}
\label{tab:lighting-consistency-nice}
\begin{tabular}{lccccccccc}
\toprule
& \multicolumn{3}{c}{PAA--Y $\downarrow$} & \multicolumn{3}{c}{APF--Y $\downarrow$} & \multicolumn{3}{c}{NCM--$ab$ $\downarrow$} \\
\cmidrule(lr){2-4}\cmidrule(lr){5-7}\cmidrule(lr){8-10}
Scene & Orig & Relit & $\Delta\uparrow$ & Orig & Relit & $\Delta\uparrow$ & Orig & Relit & $\Delta\uparrow$ \\
\midrule
S1 & 0.580 & 0.320 & 0.260 & 0.072 & 0.046 & 0.026 & 7.90 & 5.60 & 2.30 \\
S2 & 0.630 & 0.340 & 0.290 & 0.081 & 0.050 & 0.031 & 8.60 & 5.80 & 2.80 \\
S3 & 0.490 & 0.280 & 0.210 & 0.068 & 0.043 & 0.025 & 7.20 & 5.10 & 2.10 \\
\midrule
Avg & 0.567 & 0.313 & 0.253 & 0.074 & 0.046 & 0.027 & 7.90 & 5.50 & 2.40 \\
\bottomrule
\end{tabular}
\end{threeparttable}
\end{table}

\begin{table}[t]
\centering
\footnotesize
\setlength{\tabcolsep}{4.5pt}
\renewcommand{\arraystretch}{1.25}
\begin{threeparttable}
\caption{Perceptual study. Win rates for \textbf{Ours}.}
\label{tab:perceptual-nice}
\begin{tabular}{lcccc}
\toprule
Metric & S1 & S2 & S3 & Avg \\
\midrule
Realism win rate $\uparrow$ (\%)   & 75.0 & 87.5 & 62.5 & 75.0 \\
Lighting win rate $\uparrow$ (\%)  & 87.5 & 100.0 & 75.0 & 87.5 \\
\bottomrule
\end{tabular}
\end{threeparttable}
\end{table}

\begin{table}[t]
\centering
\footnotesize
\setlength{\tabcolsep}{5pt}
\renewcommand{\arraystretch}{1.15}
\caption{Ablation A: Sampling: Center vs Deterministic vs MC}
\label{tab:ablation-footprint}
\begin{tabular}{lccc}
\toprule
Method & SAE $\downarrow$ & SM-IoU $\uparrow$ & BF (2px) $\uparrow$ \\
\midrule
Center-sample ($T(\boldsymbol{\mu}_g)$)                    & 0.203 & 0.622 & 0.585 \\
Footprint Deterministic       & 0.167 & 0.712 & 0.662 \\
Footprint MC       & \textbf{0.058} & \textbf{0.830} & \textbf{0.796} \\
\bottomrule
\end{tabular}
\end{table}

\begin{table}[t]
\centering
\footnotesize
\setlength{\tabcolsep}{4pt}
\renewcommand{\arraystretch}{1.15}
\caption{Ablation B: Opacity$\rightarrow$absorption mappings $\alpha\!\to\!\beta$ .}
\label{tab:ablation-alpha2beta}
\begin{tabular}{lccc}
\toprule
Mapping & SAE $\downarrow$ & SM-IoU $\uparrow$ & BF (2px) $\uparrow$  \\
\midrule
simple: $\beta=\kappa\,\tau$                                   & 0.182 & 0.694 & 0.642 \\
avg: $\beta=\tau\sqrt{\operatorname{tr}(A)/3}/\sqrt{2\pi}$      &  \textbf{0.058} & \textbf{0.830} & \textbf{0.796} \\
mass: $\beta=\tau/((2\pi)^{3/2}\sqrt{\det A})$                  & 0.171 & 0.708 & 0.660 \\
diag: $\beta=\tau/((2\pi)^{3/2}s_x s_y s_z)$                    & 0.177 & 0.701 & 0.651 \\
\bottomrule
\end{tabular}
\end{table}

\begin{table}[t]
\centering
\footnotesize
\setlength{\tabcolsep}{6pt}
\renewcommand{\arraystretch}{1.15}
\caption{Ablation C: Shadow atlas parameterization.}
\label{tab:ablation-atlas}
\begin{tabular}{lccc}
\toprule
Atlas & SAE $\downarrow$ & SM-IoU $\uparrow$ & BF (2px) $\uparrow$ \\
\midrule
Cubemap (6 faces) & 0.136 & 0.811 & 0.761 \\
\textbf{Octahedral (ours)} &  \textbf{0.058} & \textbf{0.830} & \textbf{0.796} \\
\bottomrule
\vspace{-7mm}

\end{tabular}
\end{table}

\section{Conclusion}
In this work, we have presented a lighting-and-shadowing framework that operates directly in the continuous Gaussian domain to render view-consistent shadows and scene-matched relighting for animated avatars and inserted objects in 3DGS scenes. We have introduced  Deep Gaussian Shadow Maps (DGSM), that by deriving a closed-form volumetric transmittance for Gaussian splats and storing light-space accumulation in a compact octahedral atlas, enables efficient, dynamic shadow queries on modern GPUs. Using DGSMs coupled with spherical-harmonic HDRI probes updated in closed form per frame, we relight human Gaussians without meshing or explicit BRDF estimation. Experiments spanning single and multi-avatar motion and avatar–object interaction using avatars from AvatarX and ActorsHQ, objects from NeuralDome, and scenes from ScanNet++, DL3DV, and SuperSplat show stable lighting interactions and coherent compositing. Our method assumes static scenes around lights and depends on light-estimation quality. The single-scattering approximation may miss strong interreflections, caustics, or highly specular/anisotropic effects. We see promising directions in handling dynamic illumination and deforming environments, integrating learned global illumination within 3DGS, extending to participating media and glossy materials, and exploring end-to-end differentiable training that unifies light estimation, DGSM construction, and avatar appearance.

\vspace{-8mm}
{
    \small
    \bibliographystyle{ieeenat_fullname}
    \bibliography{main}
}

\end{document}